\documentclass[5p,authoryear]{elsarticle}

\newenvironment{widetable}[1][tbp]{\begin{table*}[#1]}{\end{table*}}
\newenvironment{widefigure}[1][tbp]{\begin{figure*}[#1]}{\end{figure*}}

\usepackage{amsmath,amssymb}
\usepackage{booktabs}
\usepackage{tabularx}
\usepackage{array}
\usepackage{multirow}
\usepackage{xcolor}
\usepackage{url}
\usepackage[hidelinks]{hyperref}
\usepackage{graphicx}
\usepackage{tikz}
\usetikzlibrary{positioning,arrows.meta}
\definecolor{nixink}{HTML}{222121}
\definecolor{nixpaper}{HTML}{F6F4F0}
\definecolor{nixbgtwo}{HTML}{EAE9E6}
\definecolor{nixaccent}{HTML}{BFD1FF}
\newcommand{\rolefont}{\fontsize{6}{7.5}\selectfont\ttfamily}
\usepackage{listings}
\lstdefinestyle{panelworkflow}{
  language=Python,
  basicstyle=\footnotesize\ttfamily,
  keywordstyle=\bfseries,
  commentstyle=\itshape\color{black!60},
  stringstyle=\ttfamily,
  showstringspaces=false,
  columns=fullflexible,
  keepspaces=true,
  frame=single,
  framesep=4pt,
  xleftmargin=6pt,
  xrightmargin=6pt,
  breaklines=true,
}
\newcolumntype{Y}{>{\raggedright\arraybackslash}X}

\journal{International Journal of Forecasting}
\biboptions{authoryear,round}

\newcommand{\artifactdoi}{10.6084/m9.figshare.33399445}
\newcommand{\artifacturl}{\href{https://doi.org/\artifactdoi}{\nolinkurl{https://doi.org/\artifactdoi}}}

\begin{document}
\begin{frontmatter}

\title{The Nixtlaverse: An Open-Source Ecosystem for Forecasting\tnoteref{repo}}
\tnotetext[repo]{Open-source project repositories: \url{https://github.com/Nixtla}.}

\author[aff1]{Olivier Sprangers\corref{cor1}}
\ead{olivier@nixtla.io}

\author[aff1]{Max Mergenthaler Canseco}
\ead{max@nixtla.io}

\author[aff1]{Marco Peixeiro}
\ead{marco@nixtla.io}

\author[aff1]{Saul Caballero Ramirez}
\ead{saul@nixtla.io}

\author[aff1]{Mariana Menchero Garc\'ia}
\ead{mariana@nixtla.io}

\author[aff1]{Jing-Qiang Goh}
\ead{jq@nixtla.io}

\author[aff1]{Han Wang}
\ead{han@nixtla.io}

\author[aff1]{Nikhil Gupta}
\ead{nikhil@nixtla.io}

\author[aff1]{Rogelio Melo}
\ead{rogelio@nixtla.io}

\author[aff1]{Senbong Gee}
\ead{senbong@nixtla.io}

\author[aff1]{Cristian Challu}
\ead{cristian@nixtla.io}

\cortext[cor1]{Corresponding author.}

\affiliation[aff1]{organization={Nixtla},
  city={San Francisco},
  state={CA},
  country={United States}}

\begin{abstract}
Large forecasting applications often combine statistical, machine-learning, and neural models. These model families solve the same problem, but they differ in their fitted state, their training procedures, and the way they parallelize work. Forecasting software must therefore either hide these differences behind a single estimator interface, or keep the families in separate packages, which forces users to rewrite data preparation and evaluation for every package.

In this paper, we present the Nixtlaverse, an ecosystem of open-source Python libraries for time series forecasting, as a case study of a third design: all libraries share the same long-format panel data and keyed forecast outputs, while every model family keeps its own specialized implementation. We demonstrate the behavior and benefits of this design through three use cases on the public M5 competition data. First, we evaluate statistical, machine-learning, and neural models, as well as an external engine from a separate ecosystem, in a single rolling-origin evaluation. The shared outputs make per-series and hierarchy-weighted accuracy metrics cheap to compute side-by-side. Second, we profile runtime and peak memory from 100 to 30,490 series and locate the bottleneck of each family: statistical fitting scales approximately linearly in the number of series, feature construction dominates machine-learning memory, and neural training time is nearly independent of panel size under a fixed training budget. Third, we reconcile the forecasts of multiple engines, including the external one, over all 42,840 series of the M5 hierarchy, using sparse reconciliation where dense implementations exhausted the memory of our machine.

These use cases establish the costs and boundaries of the design and demonstrate the utility of shared data and output contracts for cross-family evaluation and reconciliation. The success of these design choices, and by extension the Nixtlaverse itself, has been validated by the substantial public distribution, scholarly reuse, and adoption through other forecasting frameworks. The Nixtlaverse is released under permissive open-source licenses, with public datasets, reproducible examples, and verifiable benchmark artifacts. 

\end{abstract}


\end{frontmatter}

\section{Introduction}
\label{sec:introduction}

Large forecasting applications require more than fitting a model: practitioners prepare temporal panels, generate multi-step forecasts, evaluate methods over historical origins, and often aggregate forecasts for downstream decisions, frequently combining local statistical methods, global regressors, and neural models. These families solve the same problem but differ in fitted state, feature construction, training loops, preferred hardware, and natural unit of parallelism. Software can expose them through one estimator abstraction, which must accommodate those differing semantics, or through independent packages, which forces users to rewrite data preparation and evaluation. We study a third design that shares panel and forecast-output contracts while retaining specialized model-family implementations.

We state three design commitments. First, shared panel and output contracts can support cross-family evaluation even when model construction and behavioral defaults remain engine-specific. Second, scalability should be implemented within each model family because its bottleneck and natural unit of parallelism are properties of the method. Third, downstream operations should consume keyed forecasts rather than fitted models, so that evaluation and reconciliation remain independent of the producing engine. The first commitment applies a general interface principle to forecasting by
standardizing a narrow data contract while leaving implementations free behind it. What makes the contract forecasting-specific is its explicit representation of series, time, and forecast origin.

\begin{sloppypar}
The Nixtlaverse is a collection of interoperable forecasting libraries composed of \texttt{StatsForecast}, \texttt{MLForecast}, \texttt{NeuralForecast}, \texttt{CoreForecast}, \texttt{UtilsForecast}, \texttt{HierarchicalForecast}, and \texttt{DatasetsForecast}. Together, they implement these commitments through a shared long-panel representation and keyed forecast dataframes.\footnote{Throughout the paper, software packages are set in typewriter type (e.g., \texttt{StatsForecast}, \texttt{pandas}), whereas models and algorithms are set in roman type (e.g., SeasonalNaive, LightGBM).} \texttt{StatsForecast} \citep{garza2022statsforecast}, \texttt{MLForecast}, and \texttt{NeuralForecast} \citep{olivares2022neuralforecast} provide specialized forecasting engines; \texttt{CoreForecast} supplies grouped numerical operations; and \texttt{UtilsForecast} and \texttt{HierarchicalForecast}
\citep{olivares2024hierarchicalforecast} consume forecast outputs for evaluation and reconciliation.
\end{sloppypar}

We evaluate these commitments through three concrete use cases on the public M5 dataset \citep{makridakis2022m5accuracy}. Each represents a task faced by forecasting practitioners and examines the behavior and boundaries of one commitment. The first applies a common rolling-origin evaluation across all three model families and an engine external to the ecosystem (Section~\ref{sec:evaluation}). The second profiles resource use as a workflow scales from a 100-series pilot to the complete 30,490-series panel (Section~\ref{sec:scaling}). The third performs hierarchical reconciliation over all 42,840 series in the M5 hierarchy, independently of the engines that
produced the base forecasts (Section~\ref{sec:hierarchy}). The complete experimental protocol, including datasets, method configurations, and measurement procedures, is provided in \ref{app:protocol}.

The contributions of this work are:
\begin{enumerate}
  \item a comprehensive model-family boundary that specifies what the shared panel, input and output contracts guarantee, with functional semantics (e.g., training, forecasting, reconciling) deliberately remaining engine-specific;
  \item measurements, organized as an evaluation use case (Section~\ref{sec:evaluation}) and a scaling use case (Section~\ref{sec:scaling}), connecting that boundary to common rolling-origin evaluation across heterogeneous engines and to the distinct runtime and memory bottlenecks of local statistical, global machine-learning, and global neural methods;
  \item a hierarchical use case (Section~\ref{sec:hierarchy}) demonstrating that output-level interoperability supports common metrics and sparse hierarchical reconciliation while exposing differences between per-series, weighted, and coherent forecasts, and the memory boundary at which the tested dense reconcilers stop being feasible. In addition, as a direct test of extensibility, we tested an engine external to the ecosystem entering both the evaluation and reconciliation paths through the same keys; and
  \item a descriptive contribution: a license and public-infrastructure inventory of the ecosystem, and an account of the release-coordination and reproducibility obligations its multi-repository design implies.
\end{enumerate}

\section{Related Work}
\label{sec:related}

Open-source forecasting frameworks differ in which parts of the forecasting workflow they standardize. The R \texttt{forecast} package established automatic local modeling \citep{hyndman2008forecast}, while \texttt{tidyverts} combines indexed \texttt{tsibble} data with model and forecast tables \citep{wang2020tsibble,oharawild2026fabletools,oharawild2026fable}. In Python, \texttt{Prophet} exposes a single decomposable model through a small interface \citep{taylor2018forecasting}; \texttt{GluonTS} focuses on probabilistic and neural components \citep{alexandrov2020gluonts}; \texttt{Darts} exposes multiple model families through one interface \citep{herzen2022darts}; \texttt{sktime} uses a common forecaster interface with capability tags \citep{loning2019sktime}; and AutoGluon--TimeSeries places statistical, tree-based, and neural models behind a single automated selection interface \citep{shchur2023autogluon}. Pretrained forecasting models provide another workflow in which the target panel need not be used for training \citep{ansari2024chronos,das2024timesfm}; and have to be considered outside the scope of this work.

Statistical, machine-learning and neural forecasting methods are not interchangeable implementations of one identical abstraction. \citet{januschowski2020criteria} note that the distinction between the aforementioned model families is mostly of tribal nature, and discuss objective and subjective dimensions to categorize forecasting methods, such as local versus global parameterization, or complexity. These dimensions cut across the boundaries a common estimator interface must hide.  \citet[\S2.7.10]{petropoulos2022theory} document the distinct data, computational, and tuning requirements the machine-learning and neural families impose relative to statistical methods. The local--global boundary is supported by both theory and empirical evidence. \citet{monteromanso2021principles} establish the conditions under which a single global function fitted across a panel can match or outperform per-series models, while \citet{semenoglou2021crosslearning} examine the behavior of such cross-learning empirically. Because this boundary is a property of the method, Section~\ref{sec:nixtlaverse} treats the unit of parallelism as a property of the method as well. Evaluation practice for global and deep models likewise requires care over training budgets, input windows, and comparability of tuning effort \citep[\S\S 4.2.5, 4.5, 5.9]{hewamalage2021rnn}; our fixed-budget configurations are motivated by that guidance and inherit its limitations. Our contribution is not to show that these families differ, but to measure where the difference surfaces in software boundaries, and to enable the community to repeat that measurement through the released artifact.

The forecasting frameworks discussed above differ along two independent design axes, one concerning data representation and the other model exposure. A dedicated temporal object such as a \texttt{tsibble} can validate frequency and alignment when constructed, while a generic dataframe integrates directly with
existing data systems but leaves those checks to the user. Similarly, a common estimator interface presents every model family uniformly, whereas separate engines preserve the training and scaling behavior of each family. The Nixtlaverse uses the general dataframe on the first axis and separate engines on the second, and recovers what is shared, such as evaluation and reconciliation, at the level of keyed outputs instead. Treating the input of the different engines as a generic dataframe abstraction brings positive outcomes like enabling flexibility around different instances of that abstraction, such as \texttt{pandas}, Dask, or even Spark dataframes. The consequences of this design choice become visible in Section~\ref{sec:evaluation}, where behavioral defaults differ across engines and the forecast horizon is specified per call in some engines and per model in
others.

Rolling-origin evaluation measures performance over multiple historical origins \citep{tashman2000outofsample}, with special care required when target-derived features are recreated \citep{bergmeir2018note}. Curated multi-dataset archives such as that of \citet{godahewa2021monash} standardize the data side of benchmarking; the boundary we examine is the software side that consumes such data. Hierarchical reconciliation is similarly downstream of model fitting: it adjusts base forecasts to satisfy aggregation constraints \citep{hyndman2011optimal,wickramasuriya2019mint}; \citet{athanasopoulos2024reconciliation} survey the field, including the probabilistic extensions our point-forecast experiments do not reach. Frameworks also differ in where reconciliation attaches. \texttt{fabletools} attaches reconciliation to a table of fitted \emph{models}, producing coherent forecasts when the table is forecast \citep{oharawild2026fabletools}, which requires participating engines to live inside the framework. The design studied here instead reconciles keyed \emph{forecasts}, which brings independence from the producing engine. We use these tasks to test whether one output representation can support common evaluation and reconciliation across model families. \texttt{FoReco} \citep{girolimetto2026foreco} likewise reconciles base forecasts independently of the producing engine, supplied as positionally aligned numeric matrices; the design studied here differs in identifying series by explicit keys in the engines' native output format rather than by matrix position.

The prior descriptions of the individual libraries \citep{garza2022statsforecast,olivares2022neuralforecast,olivares2024hierarchicalforecast} document each package's models and interface. What they do not contain, and this paper adds, is the ecosystem-level analysis, including an explicit statement of the shared data and output contracts and their boundary, phase-level runtime and memory measurements across panel sizes for all three families under one protocol, cross-engine reconciliation over the complete M5 hierarchy with its feasibility limits, an interoperability test with an engine external to the ecosystem, and a versioned, verifiable benchmark artifact.

\section{The Nixtlaverse}
\label{sec:nixtlaverse}

\subsection{Design requirements}

Local statistical methods learn separate state per series and parallelize across a panel; global machine-learning methods materialize lag and calendar features before fitting a shared regressor; neural methods train from batches of temporal windows, often on an accelerator such as a graphics processing unit (GPU). A common workflow must therefore preserve series, timestamps, covariates, forecast origins, observations, and predictions without requiring these families to share fitted state or a scaling strategy. Historically, forecasting software was designed around automatic modeling of one series at a time \citep{hyndman2008forecast}, and forecasting across large panels was recognized as a software problem in its own right only later \citep{taylor2018forecasting}; we therefore treat scalability as an explicit, per-family design requirement.

\subsection{Panel representation}

The Nixtlaverse libraries represent a temporal panel as a long dataframe with minimum columns \texttt{(unique\_id, ds, y)} for the series, timestamp, and target. Additional columns are optional and might contain static or time-varying covariates. This format supports unequal history lengths and partitioning with common dataframe and distributed-computing systems.

Explicit series and time keys avoid dependence on row position or a rectangular array and let forecasts be joined directly to observations. A general dataframe cannot, however, infer whether a covariate is available beyond the forecast origin, or prevent every alignment error; the libraries therefore require model-specific declarations for static and future covariates. The trade-off keeps the panel simple while making assumptions explicit.

\subsection{The contract}
\label{sec:contractspec}

The shared contract is the boundary this paper examines, at the versions in Table~\ref{tab:opensource}.

\emph{Input.} The shared input is a Python dataframe, such as
\texttt{pandas}, \texttt{polars}, Spark, or Dask, containing a series identifier, a timestamp or integer time column, and a numeric target. Any additional columns represent covariates, whose static or future status is declared through the corresponding engine rather than inferred from the dataframe. \texttt{UtilsForecast} provides the common validation of frame type, required columns, and time and target dtypes, which is the only validation the contract guarantees, while each engine remains responsible for any additional formal or semantic validation and may enforce it to a different extent. Maintaining a consistent contract therefore requires engine-independent checks to be consolidated in \texttt{UtilsForecast}, although the separate-library design can allow that convergence to be deferred (Section~\ref{sec:coordination}).

\emph{Output.} A dataframe keyed by series and timestamp (plus the forecast origin under rolling-origin evaluation), with one numeric column per model and $h$ rows per series and origin, the identifiers drawn from the input panel and the timestamps continuing it at the declared frequency.

\emph{Conformance.} An external engine is a conforming producer for the downstream components if it emits exactly the output frame above; the external-engine test of Section~\ref{sec:evaluation} is the checklist executed (setup in \ref{app:external}). In principle, evaluation requires nothing further, however, specific evaluation scenarios may require additional information. For example, some reconcilers from \texttt{HierarchicalForecast} may require an in-sample frame of the same shape (series, timestamp, observed target, and fitted value per model), and evaluating with some loss functions from \texttt{UtilsForecast} requires distributional outputs or a seasonal period. Thus, beyond the output frame itself, what a producer must supply depends on the furthest downstream component it feeds. 

\emph{Evolution.} The contract serves as the ecosystem's public interface and remains unchanged across the release range examined here, while its use by six of the seven packages means that any modification to the required columns or key semantics constitutes a breaking change that the multi-repository design must coordinate manually, as discussed in Section~\ref{sec:coordination}.

\begin{widetable}[tbp]
\centering
\caption{Open-source components used in the paper. Reported package versions and their release licenses were recorded together with the public project assets in July 2026 from the projects' public repositories (URL on the title page). ``Guide'' denotes a repository-level contribution guide. The artifact records the version-pinned dependency closure used by the benchmark.}
\label{tab:opensource}
\small
\begin{tabularx}{\textwidth}{@{}l l l X@{}}
\toprule
Component & Version & Licence & Public project assets \\
\midrule
\texttt{StatsForecast}        & 2.0.3  & Apache-2.0 & source, documentation, tests/CI, guide \\
\texttt{MLForecast}           & 1.0.31 & Apache-2.0 & source, documentation, tests/CI, guide \\
\texttt{NeuralForecast}       & 3.2.0  & Apache-2.0 & source, documentation, tests/CI, guide \\
\texttt{HierarchicalForecast} & 1.3.1  & Apache-2.0 & source, documentation, tests/CI, guide \\
\texttt{CoreForecast}         & 0.0.18 & Apache-2.0 & source, documentation, tests/CI \\
\texttt{UtilsForecast}        & 0.2.16 & Apache-2.0 & source, documentation, tests/CI, guide \\
\texttt{DatasetsForecast}     & 1.0.0  & MIT        & source and dataset loaders \\
\bottomrule
\end{tabularx}
\end{widetable}

\subsection{Nixtla's forecasting libraries}

\texttt{StatsForecast} implements local statistical and econometric models using compiled numerical routines and parallel or distributed execution over series \citep{garza2022statsforecast}; the automatic exponential-smoothing model we use follows the state-space selection framework of \citet{hyndman2002statespace}. \texttt{MLForecast} creates temporal features for general regressors and manages recursive or direct multi-step prediction. \texttt{NeuralForecast} samples windows for global neural architectures and supports point and probabilistic losses \citep{olivares2022neuralforecast}. Their constructors remain family-specific, but each accepts a long panel and returns a keyed dataframe with one column per model.

\subsection{Shared components}

Figure~\ref{fig:architecture} provides an overview of the ecosystem. Datasets and long panel data enter the forecasting libraries from the left. The libraries return keyed forecast dataframes, which are consumed by evaluation and hierarchical reconciliation on the right.

\begin{widefigure}[tbp]
\centering
\resizebox{\linewidth}{!}{%
\begin{tikzpicture}[
  font=\footnotesize\sffamily,
  node distance=3.5mm and 9mm,
  box/.style={draw=nixink, line width=0.5pt, sharp corners, align=center,
              inner sep=5pt, minimum height=9.5mm, fill=nixpaper},
  engine/.style={box, fill=nixaccent, minimum width=34mm},
  shared/.style={box, fill=nixbgtwo, minimum width=36mm},
  contract/.style={box, fill=nixpaper, minimum width=32mm, rounded corners=2.5mm},
  infra/.style={box, fill=nixink, text=nixpaper},
  legend/.style={draw=nixink, line width=0.5pt, align=center, inner sep=3pt,
                 minimum height=5.5mm, font=\scriptsize\sffamily},
  arrow/.style={-{Stealth[length=1.8mm]}, nixink, line width=0.5pt},
  uses/.style={nixink!55, densely dashed, line width=0.5pt},
]
\node[contract] (panel) {long panel\\[-0.5pt]\texttt{\scriptsize(unique\_id, ds, y)}};
\node[shared, above=5mm of panel] (data) {\texttt{DatasetsForecast}\\[-0.5pt]{\rolefont DATASET LOADERS}};
\node[shared, below=5mm of panel] (prep) {\texttt{UtilsForecast}\\[-0.5pt]{\rolefont PRE-PROCESSING}};
\node[engine, right=of panel] (mf) {\texttt{MLForecast}\\[-0.5pt]{\rolefont MACHINE LEARNING}};
\node[engine, above=of mf] (sf) {\texttt{StatsForecast}\\[-0.5pt]{\rolefont STATISTICAL}};
\node[engine, below=of mf] (nf) {\texttt{NeuralForecast}\\[-0.5pt]{\rolefont NEURAL}};
\node[contract, right=of mf] (fcst) {keyed forecasts\\[-0.5pt]{\scriptsize keys \texttt{(unique\_id, ds, cutoff)},}\\[-0.5pt]{\scriptsize one column per model}};
\node[shared, above right=1mm and 8mm of fcst.east] (utils) {\texttt{UtilsForecast}\\[-0.5pt]{\rolefont EVALUATION}};
\node[shared, below right=1mm and 8mm of fcst.east] (hier) {\texttt{HierarchicalForecast}\\[-0.5pt]{\rolefont RECONCILIATION}};
\path (panel.south) -- (hier.south) coordinate[midway] (mid);
\node[infra, minimum width=128mm] (core) at (mid |- prep.south) [yshift=-6mm]
  {\texttt{CoreForecast}\quad{\rolefont COMPILED GROUPED OPERATIONS OVER VARIABLE-LENGTH SERIES}};
\draw[arrow] (data) -- (panel);
\draw[arrow] (prep) -- (panel);
\draw[arrow] (panel.east) -- (sf.west);
\draw[arrow] (panel.east) -- (mf.west);
\draw[arrow] (panel.east) -- (nf.west);
\draw[arrow] (sf.east) -- (fcst.west);
\draw[arrow] (mf.east) -- (fcst.west);
\draw[arrow] (nf.east) -- (fcst.west);
\draw[arrow] (fcst.east) -- (utils.west);
\draw[arrow] (fcst.east) -- (hier.west);
\draw[uses] (nf.south) -- (nf.south |- core.north);
\node[legend, fill=nixaccent, anchor=west] (lega) at ([xshift=10mm, yshift=-7.5mm]core.south west) {forecasting engine};
\node[legend, fill=nixbgtwo, right=3mm of lega] (legb) {shared package};
\node[legend, fill=nixpaper, rounded corners=1.8mm, right=3mm of legb] (legc) {data contract};
\node[legend, fill=nixink, text=nixpaper, right=3mm of legc] (legd) {compiled core};
\end{tikzpicture}
}
\caption{Overview of the open-source Nixtlaverse. Packages are drawn as sharp-cornered boxes and the shared data contracts as rounded boxes; the legend gives the meaning of the fills. The forecasting engines consume the same long panel representation, optionally pre-processed by \texttt{UtilsForecast} (e.g.\ filling date gaps), and return keyed forecast dataframes, which downstream evaluation and reconciliation consume regardless of the producing engine. The dashed line indicates that the three engines build on the compiled grouped operations of \texttt{CoreForecast}.}
\label{fig:architecture}
\end{widefigure}

\texttt{UtilsForecast} evaluates keyed outputs and supplies pre-processing utilities for the long panel, most importantly \texttt{fill\_gaps}, which inserts the rows needed to make every series contiguous between its first and last observation, an assumption the engines make of the input panel. These utilities, together with its plotting helpers, operate on any conforming panel and are usable independently of the forecasting libraries. \texttt{DatasetsForecast} supplies benchmark loaders, and \texttt{HierarchicalForecast} provides forecast reconciliation strategies as a post-processing step \citep{olivares2024hierarchicalforecast}. \texttt{CoreForecast} provides compiled grouped operations that the three forecasting engines build on.

\subsection{Scaling strategies}

The modular architecture reflects how the methods scale. \texttt{StatsForecast} partitions independent series fits and uses compiled routines inside each, so the dominant work grows with the number of series and the complexity of the local model. \texttt{MLForecast} constructs features over the complete panel before fitting a shared estimator, and either stage can dominate depending on the regressor. \texttt{NeuralForecast} samples and batches temporal windows, so its cost depends on the number and length of windows, the architecture, and accelerator utilization. \texttt{HierarchicalForecast} introduces cross-series matrix operations that are typically not independent across series, because the aggregation structure connects them.

The same differences determine how work is distributed once a panel no longer fits one machine. Because every row carries its keys, the panel can be partitioned without positional structure: \texttt{StatsForecast} partitions series fits through Spark \citep{zaharia2016spark}, Dask \citep{rocklin2015dask}, or Ray \citep{moritz2018ray}, dispatching to whichever backend holds the dataframe through the engine-agnostic \texttt{Fugue} layer \citep{wang2022fugue}; \texttt{MLForecast} distributes feature construction but also needs a distributed estimator; \texttt{NeuralForecast} instead supports accelerator, multi-device, and Spark-based multi-node training of one global model. Because these strategies are operationally distinct despite similar workflow methods, we measure preparation, fitting, prediction, evaluation, and reconciliation separately rather than treating one parallelization parameter as a common measure of scalability.

\subsection{Open-source implementation and coordination}
\label{sec:coordination}

Separate libraries allow users of statistical models to avoid installing a deep-learning stack, preserve model-specific choices in the public interfaces, and enable shared numerical and evaluation components to be reused without placing every model family on the same release cycle.
Table~\ref{tab:opensource} documents the open-source foundations of the case study.

The multi-repository design creates a corresponding coordination burden because data conventions, frequency handling, defaults, output schemas, and dependency versions must be aligned manually across releases, while cross-library tests cover only the combinations they exercise. A unified framework, such as the estimator-interface designs discussed in Section~\ref{sec:related}, centralizes this coordination but also couples every model family to a common dependency stack and release cadence; the design studied here instead accepts manual coordination to preserve their independence.

\subsection{Observable reach and downstream adoption}
\label{sec:impact}

The public infrastructure described above also provides partial evidence about the ecosystem's reach. To assess the impact of the Nixtlaverse, we measured several indicators: repository stars, package downloads, citations, and public case studies. We report these signals separately and attach each measurement to a fixed snapshot date.

\begin{widetable}[tbp]
\centering
\caption{Public indicators of reach for the principal Nixtlaverse libraries, retrieved on 21 August 2026. Downloads are distribution events recorded by PyPI during the preceding 30 days. Citation counts are those displayed for the canonical software records in Google Scholar; \texttt{MLForecast} did not have a separately indexed software record. These indicators are not necessarily counts of distinct users, deployments, or citing works across the ecosystem.}
\label{tab:impact}
\small
\begin{tabular}{@{}lrrr@{}}
\toprule
Library & GitHub stars & PyPI downloads & Google Scholar citations \\
\midrule
\texttt{StatsForecast}        & 4,876 & 1,942,775 & 135 \\
\texttt{MLForecast}           & 1,269 &   505,473 & --- \\
\texttt{NeuralForecast}       & 4,248 &   285,120 & 146 \\
\texttt{HierarchicalForecast} &   756 &   289,221 & 29 \\
\bottomrule
\end{tabular}
\end{widetable}

Table~\ref{tab:impact} reports public indicators for the four principal user-facing libraries as retrieved on 21 August 2026. The download counts should not be interpreted as three million distinct users or summed into an ecosystem-wide adoption figure. They include repeated installations, automated build and test environments, and package caches. Moreover, installations of the forecasting libraries also generate downloads of shared dependencies such as \texttt{CoreForecast} and \texttt{UtilsForecast}. The figures demonstrate sustained distribution activity, not the number of people or organizations using the software.

Reuse outside the Nixtlaverse provides a more structural form of adoption. At least four independently maintained forecasting projects expose Nixtlaverse implementations or data adapters. \texttt{Darts} \citep{herzen2022darts} provides a general wrapper around \texttt{StatsForecast} models and an interface for \texttt{NeuralForecast} models. \texttt{sktime} \citep{loning2019sktime} supplies adapters for several \texttt{StatsForecast} statistical models and \texttt{NeuralForecast} architectures. \texttt{AutoGluon--TimeSeries} \citep{shchur2023autogluon} bases a family of statistical and intermittent-demand models, including AutoARIMA, ETS, CES, Theta, ADIDA, Croston and IMAPA, on \texttt{StatsForecast} implementations. The \texttt{FEV} benchmarking library \citep{shchur2025fev} separately provides adapters that convert evaluation data into the Nixtlaverse representation. These integrations place parts of the ecosystem behind external APIs and benchmark contracts maintained under separate release cycles. They therefore provide stronger evidence of downstream reuse than repository attention alone.

Embedding in independently maintained products extends beyond forecasting libraries. The Databricks Many Model Forecasting solution accelerator builds dedicated \texttt{StatsForecast}, \texttt{MLForecast}, and \texttt{NeuralForecast} pipelines \citep{databricks2026mmf}; the time-series database TDengine implements several of the statistical forecasting algorithms of its analytics component TDgpt, including ETS, Theta, and complex exponential smoothing, on \texttt{StatsForecast} \citep{taosdata2026tdgpt}; and \texttt{Lightwood}, the AutoML engine behind MindsDB, declares \texttt{StatsForecast} as a required dependency and builds its ARIMA time-series mixer on it \citep{mindsdb2026lightwood}. The AutoML platform PyCaret \citep{ali2020pycaret} exposes \texttt{StatsForecast}'s AutoARIMA as an optional engine and, like \texttt{Lightwood}, reaches it through \texttt{sktime}'s adapter, an instance of second-order reuse in which one framework's adapter becomes another framework's integration path. Such chains extend into model serving: the KServe inference platform ships an AutoGluon model server that declares \texttt{AutoGluon--TimeSeries} as a dependency and thereby carries \texttt{StatsForecast} and \texttt{MLForecast} into its dependency closure \citep{kserve2026}. These are declared dependencies and shipped integrations, verified in each project's public repository on 28 August 2026; like the indicators of Table~\ref{tab:impact}, they evidence embedding, not usage volume.

The Nixtlaverse has also entered the scholarly record. Some examples include retail demand forecasting \citep{oliveira2024retail}, electricity-load forecasting \citep{delgadofernandez2025load}, hydrological-flow forecasting \citep{muniz2026flow}, and financial-volatility forecasting \citep{souto2026nhits}. The citation counts in Table~\ref{tab:impact} also illustrate a known weakness in software citation practice: \texttt{MLForecast} had no independently indexed software record on the snapshot date, despite appearing in applied studies and software comparisons.

Independent assessments and downstream reuse provide qualitative context to the telemetry. Thoughtworks placed \texttt{MLForecast} in the ``Trial'' ring of its November 2025 Technology Radar \citep[p.~42]{thoughtworks2025radar}. It reported that \texttt{MLForecast} scaled efficiently to millions of data points and consistently outperformed comparable tools in its evaluation, and described it as a compelling choice for teams operationalizing high-volume forecasting. This is evidence from one organization rather than a representative user survey, but its assessment specifically highlights automated feature construction and distributed execution as practically valuable capabilities. A further single-source signal comes from practitioner education. The textbook \emph{Modern Time Series Forecasting with Python} \citep{joseph2024modern}, for example, uses \texttt{StatsForecast} for its single-step backtesting baselines and introduces \texttt{NeuralForecast} in its chapter on specialized deep-learning architectures.

A practitioner case study from the German data consultancy m2hycon describes selecting \texttt{StatsForecast} for a customer problem involving rare component-failure events, modeled separately by customer and component, because it exposed Croston variants, IMAPA, ADIDA, and TSB through a concise common interface \citep{windler2024statsforecast}. The report states that this enabled rapid model setup and tuning and describes the resulting approach as a robust solution for the intermittent-event problem.

Independent reuse is also visible at the framework level. \texttt{AutoGluon--TimeSeries} \citep[pp.~4, 16]{shchur2023autogluon}, developed by researchers at Amazon Web Services, relies on \texttt{StatsForecast} for its local statistical models and on \texttt{MLForecast} to construct its tabular forecasters. These components are included in \texttt{AutoGluon's} \texttt{best\_quality} preset and therefore participate in its automated model-selection and ensemble workflow. 

Taken together, the Nixtlaverse has achieved substantial public distribution, documented scholarly reuse, and adoption through other forecasting frameworks. 

\section{Use case 1: five models from four engines}
\label{sec:evaluation}

A forecasting team selecting a method for a retail demand panel faces a comparison across model families: statistical baselines, a gradient-boosted regressor, a neural architecture, and often a legacy method maintained in another ecosystem entirely. The comparison is meaningful only if every candidate is evaluated under the same protocol on the same panel. In practice, such comparisons are expensive because every framework requires its own data preparation, re-indexing, and schema translation. This use case runs that task across the three resident engines (\texttt{StatsForecast}, \texttt{MLForecast} and \texttt{NeuralForecast}) and one foreign engine, and demonstrates the first commitment: shared panel and output contracts support cross-family evaluation even when model construction and behavioral defaults remain engine-specific. The benefit of the shared contracts is that practitioners can run this comparison with one merge and one evaluation call.

The panel is the M5 competition data, consisting of daily unit sales from ten Walmart stores in three US states \citep{makridakis2022m5accuracy}, containing 30,490 bottom-level series and 47.6M observations after leading zero-sales periods are removed per series, with the competition horizon of 28 days. All methods use the same three non-overlapping forecast origins, with re-estimation at every origin. The methods are a small representative set rather than a benchmark of every model in the ecosystem. We include the Seasonal Naive and AutoETS \citep{hyndman2002statespace} from \texttt{StatsForecast}; LightGBM \citep{ke2017lightgbm} from \texttt{MLForecast}, the model class that dominated the M5 accuracy competition \citep{januschowski2022trees}, with fixed target lags and date features; and NHITS \citep{challu2023nhits} from \texttt{NeuralForecast} with a fixed context length and training budget. No method-specific hyperparameter optimization is performed; the exact configurations are in \ref{app:methods}. The external engine is Croston's method \citep{croston1972forecasting} as implemented in \texttt{sktime} 1.1.0 \citep{loning2019sktime}, an intermittent-demand method appropriate for M5 from a codebase with no shared history with the libraries under study, run in a separate virtual environment with no Nixtla package installed (\ref{app:external}). We compute MASE \citep{hyndman2006accuracy} and RMSE as unweighted means over series and origins and, at the final origin, the competition's weighted root mean squared scaled error (WRMSSE) over all 42,840 series at the twelve M5 hierarchy levels \citep{makridakis2022m5accuracy}. Metric details and the one-origin rationale for WRMSSE are in \ref{app:evaluation}.

\subsection{One panel, three engines}
\label{sec:workflow}

Figure~\ref{fig:workflow} shows the shared part of the rolling-origin workflow. Each engine receives the same long dataframe and returns forecasts keyed by series, timestamp, and origin. The outputs are joined directly and scored in one evaluation call.

\begin{widefigure}[tbp]
\begin{lstlisting}[style=panelworkflow]
from datasetsforecast.m5 import M5

from statsforecast import StatsForecast
from mlforecast import MLForecast
from neuralforecast import NeuralForecast

from statsforecast.models import AutoETS, SeasonalNaive
from lightgbm import LGBMRegressor
from neuralforecast.models import NHITS

from utilsforecast.evaluation import evaluate
from utilsforecast.losses import mase, rmse
from functools import partial

h, season, freq = 28, 7, "D"
df, *_ = M5.load(directory="data")   # long panel (unique_id, ds, y)

sf = StatsForecast(freq=freq,
    models=[AutoETS(season_length=season),
            SeasonalNaive(season_length=season)])
mf = MLForecast(freq=freq,
    models=[LGBMRegressor()],
    lags=[7, 14, 28], date_features=["dayofweek", "month"])
nf = NeuralForecast(freq=freq,
    models=[NHITS(h=h, input_size=2 * h, max_steps=1000)])

cv_sf = sf.cross_validation(
    df=df, h=h, n_windows=3, step_size=h, refit=True)
cv_ml = mf.cross_validation(
    df=df, h=h, n_windows=3, step_size=h, refit=True)
cv_nf = nf.cross_validation(
    df=df, n_windows=3, step_size=h, refit=True)

keys = ["unique_id", "ds", "cutoff"]
cv = (cv_sf.merge(cv_ml.drop(columns="y"), on=keys)
           .merge(cv_nf.drop(columns="y"), on=keys))

metrics = [partial(mase, seasonality=season), rmse]         
scores = evaluate(cv, metrics=metrics, train_df=df)
\end{lstlisting}
\caption{The M5 rolling-origin workflow: one panel, three family-specific constructors, shared cross-validation calls, a keyed merge, and one evaluation call. Seeds and logging are omitted; the executable configuration is provided in the benchmark artifact.}
\label{fig:workflow}
\end{widefigure}

Model construction remains specific to each family, with statistical models
requiring a seasonal period, the machine-learning engine requiring features and
a regressor, and the neural model requiring an architecture, context, horizon,
and training budget. To ensure that evaluation workflows are defined
consistently across engines, we recommend specifying the origin spacing and
refitting policy explicitly, as shown in Figure~\ref{fig:workflow}. The shared
data contract makes the resulting outputs composable while keeping these
behavioral differences visible.

The workflow in Figure~\ref{fig:workflow} could be executed without engine-specific reshaping or schema translation using three shared contracts, namely a long input panel keyed by series and timestamp, an explicit rolling-origin request, and forecast outputs keyed by series, timestamp, and
origin. These outputs could be joined and scored in a single evaluation call,
with the remaining code limited to dropping the duplicated observed-target
column from two frames before the join. Model-specific requirements remained confined to the constructors, including a seasonal period for statistical models,
feature definitions and a regression estimator for machine learning, and the
architecture, context length, and training budget for the neural engine. One
remaining workflow difference was how the forecast horizon was specified,
with two engines accepting it per call and the third encoding it in the model.
The contracts therefore did not make the estimators interchangeable.

Shared keys do not eliminate every coordination point because default argument
values are not synchronized across libraries, requiring both the spacing
between origins and the refitting policy to be specified explicitly, as shown
in Figure~\ref{fig:workflow}. This reflects the trade-off described in
Section~\ref{sec:nixtlaverse}, whereby the shared data contract guarantees that
outputs can be combined while each library retains responsibility for its
behavioral defaults.

We run the same workflow with an external engine, Croston's method from \texttt{sktime}, to demonstrate sufficiency of the output contract. The cost of `admission' is a 121-line adapter that reads the panel, fits the model per series and origin, and emits the keyed frame; the adapter is included in the benchmark artifact. Its forecasts then pass through the merge and evaluation call of Figure~\ref{fig:workflow} unchanged, scoring four engines from two ecosystems side by side.

\subsection{Benefits of output-level interoperability}
\label{sec:accuracy}

Table~\ref{tab:accuracy} reports the rolling-origin results on the complete panel. It is not a comparison of methods: the table contains a small, deliberately untuned selection of models, and none approaches the tuned ensembles that won the competition. The table characterizes what one merged frame makes visible about five fixed configurations, one of them produced outside the Nixtlaverse, and its value lies in the columns disagreeing.

\begin{widetable}[tbp]
\centering
\caption{Rolling-origin results on the complete M5 bottom-level panel (30{,}490 series, three origins, re-estimation at every origin), for the fixed configurations of \ref{app:methods}. Metrics use raw, unclipped engine outputs. MASE and RMSE are unweighted means over series and origins; the bracketed 95 percent bootstrap intervals resample whole series, and the corresponding RMSE intervals, which are of comparable width, are in the artifact. WRMSSE is the competition's weighted metric over all twelve hierarchy levels at the final origin. The Total column reports the aggregate forecast of the final origin relative to the observed total; cross-validation times and the resampling details are in \ref{app:evaluation}. The external Croston engine runs in an isolated environment with no Nixtla package installed (\ref{app:external}).}
\label{tab:accuracy}
\footnotesize
\begin{tabularx}{\textwidth}{@{}l Y r@{~}l r r r@{}}
\toprule
& & \multicolumn{2}{c}{MASE} & & & \\
\cmidrule(lr){3-4}
Model & Engine & mean & 95\% CI & RMSE & WRMSSE & Total \\
\midrule
SeasonalNaive & \texttt{StatsForecast} & 1.196 & [1.175, 1.229] & 1.781 & 0.847 & 0.99 \\
AutoETS       & \texttt{StatsForecast} & 1.050 & [1.028, 1.084] & 1.382 & 0.673 & 0.99 \\
LightGBM      & \texttt{MLForecast}    & 1.568 & [1.544, 1.598] & 1.534 & 1.045 & 1.06 \\
NHITS         & \texttt{NeuralForecast} & 0.874 & [0.856, 0.906] & 1.421 & 1.888 & 0.70 \\
\midrule
Croston       & \texttt{sktime (external)} & 1.073 & [1.056, 1.101] & 1.412 & 0.957 & 1.03 \\
\bottomrule
\end{tabularx}
\end{widetable}

NHITS attains the best per-series MASE but the worst hierarchical WRMSSE, while
the approximately unbiased AutoETS shows the opposite pattern. This reversal is
consistent with the evaluation objectives because absolute-error training
targets conditional medians, which are often zero for intermittent retail
demand, while an unweighted absolute-error metric can favor behavior that a
sales-weighted squared-error metric penalizes
\citep{kolassa2016countdata}. Across the three origins, NHITS predicts only
66--70\% of observed demand in aggregate and produces 11.7--16.7\% negative
forecasts, causing bottom-level errors to accumulate into downward bias at
sales-weighted upper levels. Although this pattern is consistent with its
default mean-absolute-error loss, establishing causality would require
refitting under a loss targeting a different central tendency.

The ordering is largely stable under resampling and retraining. Across models,
bootstrap intervals overlap only for AutoETS and Croston, while three additional
NHITS seeds change mean MASE by at most 0.002, keep the aggregate ratio between
0.66 and 0.73, and produce WRMSSE values between 1.732 and 1.888, still far from
the next engine. The shared output boundary makes both metrics straightforward
to compute on the same forecasts, exposing a disagreement that either metric
alone would conceal.

The disagreement also reveals a limit of the shared-output boundary. MASE and
RMSE can be computed from the merged forecast frame through a single
\texttt{UtilsForecast} evaluation call, whereas WRMSSE requires a separate
workflow that pivots the final origin, joins the official product hierarchy, and
invokes the competition evaluator. The shared schema makes the per-series and
hierarchy-weighted results directly comparable, but the latter requires more effort to obtain, while the more readily available
per-series metric ranks NHITS first. Relying on that ranking alone would select
a model whose forecasts sum to only 70\% of observed demand, a serious
limitation for inventory or revenue planning regardless of its per-series
score. Output interoperability standardizes the data exchanged between
components, but it does not determine which metric or weighting scheme should
guide model selection.

The shared frame also makes post-processing sensitivity checks straightforward.
We evaluate the raw forecasts to avoid model-specific adjustments, but clipping
negative predictions at zero changes WRMSSE only from 0.673 to 0.670 for
AutoETS and from 1.888 to 1.885 for NHITS, without altering any model ranking.
The disagreement between the metrics is therefore not an artifact of evaluating
unadjusted outputs.

\paragraph{Limitations}
This use case exercises only target lags and calendar features, not the static- and future-covariate declarations that Section~\ref{sec:nixtlaverse} identifies as the contract's error-prone edge. Because the configurations are untuned and the neural model has a fixed step budget, Table~\ref{tab:accuracy} compares equal tuning effort rather than converged or equal-exposure training and does not rank the underlying methods (\ref{app:methods} states the consequences). The external test covers one method from one engine through an adapter written by us; it establishes that the contract admits a foreign producer, not that every external engine maps this cheaply.

\section{Use case 2: scaling from a pilot to the full panel}
\label{sec:scaling}

\begin{widefigure}[tbp]
\centering
\includegraphics[width=0.62\linewidth]{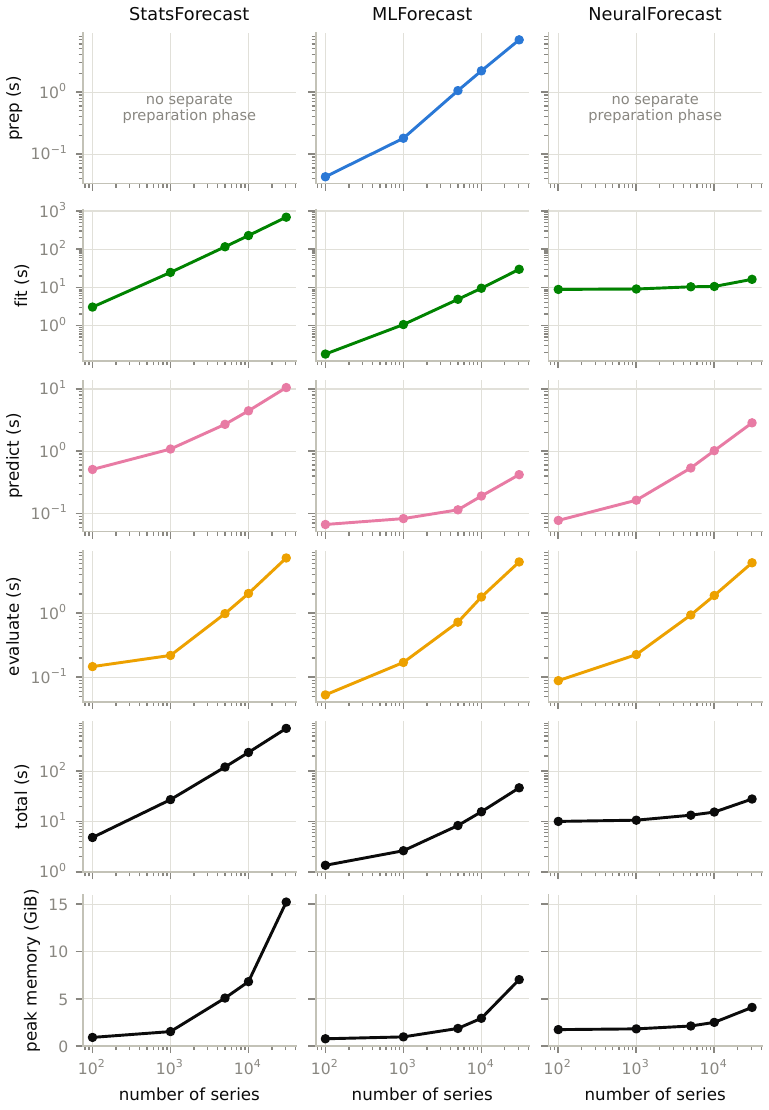}
\caption{Resource profile per engine (columns) as the number of series increases, for subsets drawn proportionally in product category and store. The first four rows show the runtime of each computational phase separately (log--log; the statistical and neural engines have no separate preparation phase), the fifth row the total runtime over all recorded phases including data loading, and the bottom row peak memory as aggregate proportional set size (PSS) during preparation, fitting, and prediction. Medians over five cold-start repetitions; axes are shared across engines within each row.}
\label{fig:scaling}
\end{widefigure}

A team moving from a pilot of one hundred series to the complete panel must
determine which computational phase will dominate, how much memory to provision,
how many workers to use, and when a distributed backend justifies its overhead.
Answering these questions requires measurements by phase and model family
because the second commitment assigns scalability to each family, making both
the bottleneck and the natural unit of parallelism properties of the underlying
method. These measurements allow practitioners to provision resources for the
phase that will dominate at their target scale.

We create nested subsets of the main panel with an increasing number of series, drawn by seeded sampling proportional in product category and store so that each subset approximates the composition of the complete panel, and reserve the final 28 days of each series as a common holdout. Data preparation, fitting, prediction, and evaluation are measured separately, recording wall-clock time and peak memory as aggregate proportional set size (PSS) over parent and worker processes; the panel-size sweep uses five measured repetitions per configuration, each in a fresh process, and we report medians. These subsets profile resources only; comparative accuracy is reported on the complete panel in Section~\ref{sec:accuracy}. The full measurement protocol, including the subset-composition rationale and the treatment of timeouts and out-of-memory kills, is in \ref{app:measurement}.

Figure~\ref{fig:scaling} separates the runtime of every engine into its computational phases as the panel grows from 100 to 30{,}490 series. All 85 runs across 17 engine--panel-size configurations completed.

The dominant phase confirms the computational structure of Section~\ref{sec:nixtlaverse}. For the statistical engine, fitting dominates at every size and scales approximately linearly with the number of series, from 3.0 seconds at 100 series to 691 seconds at the full panel. For the machine-learning engine, fitting the shared gradient-boosting estimator dominates (29.8 seconds at full scale), with feature construction adding 7.0 seconds; recursive prediction remains below one second even at the full panel because the lag updates are vectorized across series. For the neural engine, a fixed 1{,}000-step budget makes training time weakly dependent on panel size by construction (8.8 to 16.2 seconds on one GPU); this is a fixed-compute profile rather than a comparison at equal data exposure or converged training.

Machine learning has the lowest total time from 100 through 5{,}000 series; at 10{,}000 series the machine learning and the neural engine are indistinguishable (15.6 against 15.4 seconds), and at the full panel the fixed-budget neural engine is fastest (28 against 47 seconds). These totals sum every recorded phase, including the data-loading and evaluation phases no engine avoids, which together account for about nine of the neural engine's 28 seconds at full scale, so the engine-specific difference is wider than the totals suggest. The statistical engine is slowest beyond 100 series. Peak memory during preparation, fitting, and prediction is 15.2~GiB for the two-model statistical portfolio and 13.8~GiB for AutoETS alone (retained per-series fitted state), 7.0~GiB for machine learning (the materialized feature table), and 4.1~GiB for the neural engine (streamed window batches), rising to 5.6~GiB for the neural engine if evaluation is included and unchanged for the others.

\paragraph{Effect of parallelization}
Runtime growth with panel size can be summarized by a scaling exponent,
computed as the slope of fitting time against the number of series on log--log
axes, with an exponent of 1 indicating that fitting time increases in direct
proportion to the panel size. Table~\ref{tab:parallel} reports this sweep for the statistical engine (\texttt{StatsForecast} fitting AutoETS alone), measured between 100 and 1{,}000 series as medians of three repetitions. The exponent is 0.99 on one worker, 0.95 on four, and 0.84 on twenty; the speedups over one worker are 3.4 and 6.9 times at 100 series against 3.75 and 9.8 times at 1{,}000. Per-series work is thus approximately linear, as the single-worker exponent shows, and parallel execution understates it because per-worker overheads (process startup, scheduling, and possible thread contention between workers and the numerical libraries they call, whose thread counts we did not restrict) amortize over more series as the panel grows. Reporting a parallel exponent as a property of the method would conflate the two.

\begin{table}[tbp]
\centering
\caption{Effect of parallelization for the statistical engine: \texttt{StatsForecast} fitting AutoETS alone, medians of three repetitions in fresh processes. Twenty workers is every logical CPU of the machine (\ref{app:repro}). The exponent is the slope of fitting time against the number of series on log--log axes between the two panel sizes; speedups are relative to one worker at the same panel size.}
\label{tab:parallel}
\footnotesize
\begin{tabular}{@{}r rr rr r@{}}
\toprule
 & \multicolumn{2}{c}{100 series} & \multicolumn{2}{c}{1{,}000 series} & \\
\cmidrule(lr){2-3}\cmidrule(lr){4-5}
Workers & Fit (s) & Speedup & Fit (s) & Speedup & Exponent \\
\midrule
1  & 23.8 & ---          & 233.8 & ---           & 0.99 \\
4  &  7.0 & $3.4\times$  &  62.3 & $3.75\times$  & 0.95 \\
20 &  3.4 & $6.9\times$  &  23.8 & $9.8\times$   & 0.84 \\
\bottomrule
\end{tabular}
\end{table}

\paragraph{Portfolio versus single model}
Every engine accepts several models in one call; our configuration exercises this only in the statistical engine, which fits the
SeasonalNaive and AutoETS models. To keep engine comparisons single-model, we profiled AutoETS alone at the two largest sizes. Adding SeasonalNaive costs 1.6 to 2.0~GiB of retained state at the full panel (about 0.3~GiB at 10{,}000 series) and no measurable time. The apparent time difference in the block-ordered pass reversed sign under interleaved, counterbalanced repetitions, identifying it as machine-state drift rather than model count (\ref{app:measurement} reports the diagnosis). For SeasonalNaive, the cost is not the model's parameters, which are seven values per series, but the in-sample fitted values, residuals, and stored series that the fitted object retains per model, so it grows with observations rather than with series.

\paragraph{Distributed backend}
In this experiment, \texttt{StatsForecast} generated forecasts for the two-model
portfolio across the full panel using two execution modes, local multiprocessing
and the Ray backend on a local cluster, both accessed through the
\texttt{Fugue} dispatch layer described in
Section~\ref{sec:nixtlaverse}. The two backends gave numerically identical forecasts, which follows from the design: partitioning by series changes where each local model is fitted, not how. Table~\ref{tab:distributed} summarizes the single-machine cost. Ray took 897 seconds in total, of which 32 were a one-time conversion of the in-memory panel into Ray's distributed object store and 865 the forecasting pass itself, against 738 seconds for multiprocessing, and it used 19.2~GiB against 3.1~GiB, reflecting the object store and per-worker copies. We report the conversion separately because it is paid once per dataset, not per forecasting call. Ray's value lies not in improving single-machine performance but in providing a scale-out path, since the same specification accepts a distributed dataframe in place of a \texttt{pandas} dataframe without requiring changes to the modeling code \citep{mckinney2010pandas}. The streaming interface peaks at 3.1~GiB while forecasting and 3.4~GiB while loading, whereas the fit-then-predict interface used in the scaling experiment, which retains the fitted state of every series, peaks at 15.2~GiB for the same two-model portfolio. Both choices materially affect memory, through different mechanisms: the backend contrast (3.1 against 19.2~GiB) reflects Ray's object store and per-worker copies, the price of the scale-out path, while the workflow contrast (3.1 against 15.2~GiB) reflects retained per-series fitted state. Because workflow and backend were not crossed in a complete $2\times2$ design, these contrasts do not rank the two effects. Each distributed figure comes from a single run, while the 15.2~GiB result is the median of five runs.

\begin{table}[tbp]
\centering
\caption{Distributed backend against local multiprocessing for the statistical engine only: \texttt{StatsForecast} forecasting the SeasonalNaive--AutoETS portfolio over the full 30{,}490-series panel through the streaming \texttt{.forecast} interface, one measured run per backend. Construction is the one-time conversion of the in-memory panel into Ray's distributed object store. Peak memory is aggregate proportional set size over parent and worker processes.}
\label{tab:distributed}
\footnotesize
\begin{tabular}{@{}l rrr r@{}}
\toprule
 & \multicolumn{3}{c}{Runtime (s)} & Peak memory \\
\cmidrule(lr){2-4}
Backend & Forecast & Constr. & Total & (GiB) \\
\midrule
Multiprocessing & 738 & --- & 738 & 3.1 \\
Ray             & 865 & 32  & 897 & 19.2 \\
\bottomrule
\end{tabular}
\end{table}

\paragraph{Limitations}
We did not compare these family-specific strategies with a family-agnostic scaling primitive, and absolute runtimes and GPU results remain hardware- and driver-dependent. The transferable findings are the dominant phases under these configurations, the approximately linear growth of single-worker per-series fitting, and the mechanisms that would change the profiles: a cheaper regressor shifts machine learning toward feature construction, while an unfixed neural budget recouples training to panel size.

\section{Use case 3: coherent forecasts over the complete hierarchy}
\label{sec:hierarchy}

A demand planner consumes forecasts at every level of a retail hierarchy and requires them to be coherent, since a state-level total that differs from the sum of its stores creates problems for inventory and revenue planning regardless of per-series accuracy. Forecast reconciliation adjusts base forecasts to satisfy the aggregation constraints, and the third commitment says it should consume keyed forecasts rather than fitted models, so that the planner can change the producing engine, or add one from outside the ecosystem, without changing the reconciliation path. The benefit of reconciling keyed forecasts is that practitioners can swap or add base engines without modifying the reconciliation code.

The published twelve-level M5 aggregation structure contains 42,840 series and 67.6M observations. We create base forecasts with AutoETS and LightGBM, store both in the same keyed dataframe, and reconcile with fixed BottomUp and MinT configurations where feasible. The MinT configuration is \texttt{MinTraceSparse} with \texttt{method="wls\_var"}, a diagonal approximation of the forecast-error variance estimated from the in-sample residuals of each base engine (\ref{app:hierarchical} gives the full configuration). This use case runs a single 28-day holdout at the final origin rather than the three rolling origins of Section~\ref{sec:evaluation}, so its MASE values are not directly comparable with those of Table~\ref{tab:accuracy}. The aim is to demonstrate that reconciliation can be implemented independently of the base forecasting method, not that one base model performs best for every hierarchy.

Both base-forecast dataframes entered the same reconciliation implementation directly from the shared keyed representation, so output-level interoperability decoupled reconciliation from model fitting. However, the contract guarantees neither coherent base forecasts, nor improved accuracy, nor computational feasibility. Table~\ref{tab:reconciliation} reports these boundaries on the complete 42{,}840-series hierarchy. Base forecasts violate the aggregation constraints by up to 1{,}051 units for AutoETS, 18{,}071 for LightGBM and 204 for the external engine over the 28-day horizon. For these reconcilers coherence holds by construction, since upper levels are produced by summing the reconciled bottom level; \texttt{HierarchicalForecast} provides coherence utilities to verify that the emitted frames realize the constraint exactly.

\begin{widetable}[tbp]
\centering
\caption{Base and reconciled forecasts on the complete M5 hierarchy (42{,}840 series, 28-day holdout at the final origin), using sparse reconciliation implementations. MASE is the mean over series; the total level is the single top series. Runtime covers the reconciliation step only, per base engine, from one measured run each; it excludes the time of hierarchy construction. A second complete pass of the same configurations is retained in the artifact as superseded; Section~\ref{sec:hierarchy} reports how far it differs. The dense configurations attempted, BottomUp and MinT-shrink, did not complete; dense MinT with the \texttt{wls\_var} estimator was not run (Section~\ref{sec:hierarchy}). Variance-weighted reconciliation is not run for the external engine, which does not supply in-sample fitted values (\ref{app:external}).}
\label{tab:reconciliation}
\small
\begin{tabularx}{\textwidth}{@{}l Y r r r r r r@{}}
\toprule
 & & \multicolumn{3}{c}{MASE, all levels} & \multicolumn{3}{c}{MASE, total level} \\
\cmidrule(lr){3-5}\cmidrule(lr){6-8}
Forecasts & Runtime (s) & AutoETS & LightGBM & Croston & AutoETS & LightGBM & Croston \\
\midrule
Base            & ---            & 1.031 & 1.476 & 1.047 & 0.653 & 2.163 & 1.457 \\
BottomUp        & 60 / 57 / ---   & 1.029 & 1.653 & 1.047 & 0.677 & 1.723 & 1.439 \\
MinT (wls\_var) & 179 / 197 / ---& 1.033 & 1.321 & ---   & 0.672 & 0.605 & ---   \\
\bottomrule
\end{tabularx}
\end{widetable}

The MASE columns in Table~\ref{tab:reconciliation} should not be interpreted as
a ranking, but as evidence that reconciliation redistributes error rather than
consistently reducing it, with the resulting distribution determined by the
base forecasts as well as the reconciliation method. For AutoETS, whose forecasts aggregate almost without bias (Section~\ref{sec:accuracy}), every reconciler changes accuracy marginally. For LightGBM, whose total is heavily biased, bottom-up aggregation repairs the total but carries bottom-level errors upward, degrading the intermediate levels most sharply (departmental MASE moves from 1.149 to 5.694), while variance-weighted MinT redistributes the adjustment and improves every reported level. Coherence is guaranteed by construction while improvement is guaranteed only under the mathematical properties of the reconciliation method.

The reconciliation path also passes the external-engine test described in~\ref{app:external}. The foreign engine's base forecasts for all 42{,}840 series entered the same sparse bottom-up reconciler directly from the keyed frame and reproduced the pattern observed for the resident engines, with aggregation violations before reconciliation, an aggregation residual of exactly zero afterward, and only marginal changes in accuracy consistent with its nearly unbiased totals.

At the scale of M5, memory rather than runtime becomes the limiting resource. Constructing the twelve-level hierarchy took 258 seconds and peaked at 28.5~GiB before reconciliation began, while a dense summing matrix would itself require 9.7~GiB in double precision ($42{,}840 \times 30{,}490 \times 8$ bytes). Both the dense BottomUp and MinT-shrink runs were terminated by the operating system.

Sparse linear algebra removes the expensive matrix but not the data. Although the matrices used by the sparse solver are small, the aggregated panel and the 66-million-row in-sample fitted frame remain resident because the output contract passes the complete frame to the variance estimator, even though a diagonal estimator mathematically requires only one variance per series. The sparse MinT step therefore still peaked at 43.6~GiB, and was itself terminated by the operating system when both engines' intermediates were resident in one process; restructured to hold one engine at a time, it completed in 179 and 197 seconds (Table~\ref{tab:reconciliation}). 

\paragraph{Limitations}
Sparse MinT is not guaranteed to match the dense implementation of the same estimator; at this scale the dense configurations exhausted memory, so we could not bound the difference. Variance-weighted reconciliation remains undemonstrated for external producers because it additionally requires in-sample fitted values through the same keyed contract. The observed feasibility boundary is a single threshold crossing, not a general comparison of workflow or algorithm choices.

\section{Discussion}
\label{sec:discussion}

The three use cases characterize the commitments within the limits of one ecosystem, one dataset, and the versions in Table~\ref{tab:opensource}. Their value lies in separating guarantees checkable at package boundaries from behaviors that cannot be standardized away. This section synthesizes their implications; experiment-specific limitations accompany each use case, and study-wide limitations are collected below.

\paragraph{C1: Share the data contract, not the estimator}
The boundary used here is explicit series and time keys on the input, an explicit origin schedule, and series, time, and origin keys on the output. These sufficed for common evaluation across families whose fitted states have nothing in common, but not to standardize feature availability, loss functions, refitting defaults, horizon placement, or probabilistic semantics, which remain visible engine responsibilities. The evidence therefore supports a narrow interoperability claim: shared keys standardize exchange, not behavioral semantics.

\paragraph{C2: Implement scalability inside the model family}
The phase profiles of Section~\ref{sec:scaling} are consistent with this commitment, and show that partitioning by series is not a general scaling primitive. That these methods have different bottlenecks is a property of the methods, established independently \citep{januschowski2020criteria}; the use case measures where the differences surface. Memory, however, depends on the complete execution path rather than the model family alone: retaining per-series fitted state (``fit-then-predict'') raised local memory from 3.1 to 15.2~GiB relative to the streaming \texttt{.forecast} interface, while Ray's object store and per-worker copies raised streaming memory from 3.1 to 19.2~GiB. Both choices matter, and neither contrast was measured on a common design, so the commitment concerns where scaling behavior is implemented, not comparative efficiency; capacity planning must account for both model-state residency and backend overhead.

\paragraph{C3: Let downstream components consume forecasts, not models}
A single joined output table allowed both base engines to use the same reconcilers and made their evaluation metrics directly comparable, while the external test demonstrated extensibility by routing an engine from another ecosystem through the same evaluation and reconciliation workflows using a 121-line adapter (Sections~\ref{sec:evaluation} and~\ref{sec:hierarchy}). The full-hierarchy use case also establishes that schema-compatible base forecasts need not be coherent, reconciliation need not improve accuracy, and dense methods need not remain computationally tractable. Section~\ref{sec:accuracy} reveals a related boundary because the contract standardizes the data exchanged between components but not the evaluation objective, allowing the most convenient metric to produce a ranking contradicted by the hierarchy-weighted metric. These results support C3 for the point-forecast evaluation and reconciliation operations tested here.

\paragraph{Scope conditions}
These commitments are not universally advantageous. When covariate handling is a primary source of workflow errors, a dedicated time-series object that validates covariates at construction may be preferable to a generic dataframe, weakening the case for C1. When a single small team maintains the entire stack, a multi-repository architecture adds coordination costs without providing the benefit of independent release cycles. A unified estimator interface may similarly be preferable when the primary objective is automated search across
model families, weakening the case for C2. Finally, when production relies on only one model family, a shared contract has few components to connect, limiting the value of C1 and C3. The commitments are most useful when several engines evolve independently, multiple model families operate in production, and downstream workflows must support all of them.

\subsection{Limitations}

Our evidence comes from a small set of untuned methods, one dataset, one ecosystem, and fixed software versions; it neither ranks model families nor establishes general efficiency. Results depend on dataset size, seasonal patterns, covariates, hardware, and model settings. M5 in particular is daily, highly intermittent, strongly day-of-week seasonal, and hierarchical; the negative-forecast rates of Section~\ref{sec:accuracy} and the necessity of sparse reconciliation should not be expected to transfer unchanged to smoother or shallower panels.

Every engine we measure was built around the commitments we examine, so we cannot weigh them against the unified-estimator alternative of Section~\ref{sec:related}, although we expect such an interface to absorb some of the coordination costs of Section~\ref{sec:workflow}. Whether the keyed contract is minimal remains untested, as does whether every downstream operation can be expressed over keyed outputs; some operations may require fitted models rather than their outputs. We evaluate point forecasts only: probabilistic forecasting, probabilistic reconciliation in the sense of \citet{panagiotelis2023probabilistic}, and model-dependent forecast combination remain outside scope. The long-data representation is flexible but less strict than a dedicated time-series object, so errors such as incorrectly labeled future covariates cannot be detected from the dataframe alone, and the open-source inventory records visible project infrastructure at one date; it does not measure governance quality, contributor diversity, maintenance responsiveness, or long-term sustainability.

Finally, this is not an independent evaluation: the authors are not at arm's length from the software under study, and the declaration of competing interest at the end of this article states the relationship; the published protocol, raw results, and unfavorable outcomes exist so that this can be audited. Descriptions of other frameworks rest on their published documentation, and we invite their maintainers to correct any factual description.

\section{Conclusion and Future Work}
\label{sec:conclusion}

We presented the Nixtlaverse as a case study of three design commitments for open-source forecasting software: share panel and output contracts rather than fitted estimators, place scalability within model-family implementations, and define downstream operations over keyed forecasts. Its multi-repository implementation preserves specialized dependencies and interfaces, but makes compatibility testing and versioned research artifacts part of the software's coordination burden. The Nixtlaverse has achieved substantial public distribution, documented scholarly reuse, and adoption through other forecasting frameworks. 

We've demonstrated that explicit input, origin, and output keys sufficed to evaluate all three families (and one engine external to the ecosystem) on the complete M5 panel without per-engine reshaping, while leaving their constructors and behavioral semantics distinct. We offer these boundaries, rather than surface-level interface uniformity, as the transferable lesson. Whether they generalize would require replication on a second ecosystem or a domain with different intermittency and hierarchy depth, and both remain open.

Three findings do not depend on the specific libraries a reader adopts. First, capacity planning must account for both workflow and backend. For the per-series engine, retaining fitted state raised peak memory from 3.1~GiB under local streaming to 15.2~GiB under local fit-then-predict, while Ray's object store and per-worker copies raised streaming memory to 19.2~GiB. These measurements identify two material mechanisms rather than ranking their effects: deployments must budget separately for fitted-state residency and distributed-execution overhead. Second, sparse linear algebra is necessary at scale: a dense double-precision summing matrix for the M5 hierarchy requires 9.7~GiB before a single forecast is reconciled. Finally, the measurements quantify, for this workload, the property that per-series parallelization does not transfer to global models: one parallelization control across model families describes an interface rather than a scaling behavior.

The next step for the ecosystem is to reduce the coordination burden described in Section~\ref{sec:coordination} by consolidating the packages into a single workspace with a shared dependency lockfile, for example through a \texttt{uv} workspace. Cross-package changes could then be developed and tested together while each package remains separately installable, preserving the ecosystem's modularity while removing much of its manual coordination overhead.

\section*{CRediT author statement}

Olivier Sprangers: Conceptualization, Data curation, Formal analysis, Investigation, Methodology, Project administration, Software, Validation, Visualization, Writing--original draft, Writing--review and editing. Max Mergenthaler Canseco, Marco Peixeiro, Saul Caballero Ramirez, Mariana Menchero Garc\'ia, Jing-Qiang (JQ) Goh, Han Wang, Nikhil Gupta, Rogelio Melo, Senbong Gee, and Cristian Challu: Software, Writing--review and editing.

\section*{Acknowledgements}

The authors thank Jos\'e Morales, Kin Gutierrez, and Deven Mistry.

\section*{Funding}

This research did not receive any specific grant from funding agencies in the public, commercial, or not-for-profit sectors.

\section*{Declaration of competing interest}

The authors work at Nixtla, the company that develops the software examined as the principal subject of this article. This relationship may be perceived as a competing interest.

\section*{Declaration of generative AI and AI-assisted technologies in the writing process}

During the preparation of this work the authors used OpenAI Codex and Anthropic Claude Code to assist with literature discovery and drafting. The authors reviewed and edited the content and take full responsibility for the content of the article.

\section*{Data and code availability}

The accompanying artifact contains the source code, dependency specifications, integrity manifests, raw outcomes, and commands for every result, including the external-engine adapter and its isolated-environment specification (\ref{app:external}). The M5 source files are not redistributed. The artifact is publicly available at \artifacturl.

\bibliographystyle{elsarticle-harv}
\bibliography{references}

\appendix

\section{Experimental protocol}
\label{app:protocol}

This appendix specifies the datasets, method configurations, and measurement methodology behind the use cases of Sections~\ref{sec:evaluation}--\ref{sec:hierarchy}. The artifact at \artifacturl{} contains the code to reproduce.

\subsection{Datasets}
\label{app:datasets}

We use the M5 competition data: daily unit sales from ten Walmart stores in three US states \citep{makridakis2022m5accuracy}. After leading zero-sales periods are removed per series, the bottom-level panel contains 30,490 series and 47,649,940 observations. It supports the evaluation and scaling use cases. The published twelve-level aggregation structure contains 42,840 series and 67,687,697 observations and supports the hierarchical use case. We use the competition horizon of 28 days throughout; the frozen dataset manifest records the source files and hashes.

\subsection{Method configurations}
\label{app:methods}

We do not perform method-specific hyperparameter optimization, using documented defaults or one pre-specified configuration. The parameters we set are: season length 7 for SeasonalNaive and AutoETS, the latter selecting its error--trend--seasonal form automatically; target lags $\{7, 14, 28\}$ and \texttt{dayofweek} and \texttt{month} date features for LightGBM, with recursive multi-step prediction; and context length 56 ($2h$), a 1{,}000-step training budget, start padding (enabled because some M5 series are shorter than the context length at the earliest forecast origins), and the default mean-absolute-error loss for NHITS. Every engine uses horizon $h = 28$, daily frequency, and a seed fixed per repetition; all remaining parameters take their library defaults at the versions in Table~\ref{tab:opensource}.

Fixing the configurations makes tuning budget irrelevant to the comparison; the absence of tuning penalizes engines whose defaults are weak on intermittent retail data, and the fixed neural step budget insulates the neural engine from the panel-size axis that Section~\ref{sec:scaling} measures. The configurations are therefore comparable in effort, not in fitted quality, and Section~\ref{sec:accuracy} interprets the accuracy results accordingly.

\subsection{Evaluation protocol}
\label{app:evaluation}

All methods use the same three non-overlapping forecast origins, 28-day horizon, and observed target values, with a model re-estimated at every origin. We compute MASE \citep{hyndman2006accuracy} and RMSE at every origin with \texttt{UtilsForecast} from the combined forecast dataframe; the MASE denominator is the in-sample seasonal-naive mean absolute error recomputed from the data preceding each origin, with no floor imposed on it. Both metrics are unweighted means over series and origins. For the final origin, we additionally report the competition's weighted root mean squared scaled error (WRMSSE), which aggregates accuracy over all 42,840 series at the twelve M5 hierarchy levels \citep{makridakis2022m5accuracy}. WRMSSE is reported at one origin because the competition weights are defined for a specific evaluation window, so the weighted comparison in Section~\ref{sec:accuracy} is established at that origin and not across the three; the aggregate forecast ratios reported alongside it do not depend on the competition weights and are given at all three origins.

The intervals in Table~\ref{tab:accuracy} are percentile bootstrap intervals over 2,000 resamples of whole series. The intervals condition on one training realization of the neural model; the seed sensitivity reported in Section~\ref{sec:accuracy} comes from retraining NHITS under three further seeds at full scale, recorded in the artifact. 

\subsection{Resource-measurement protocol}
\label{app:measurement}

For the scaling use case, we create nested subsets of the main panel with an increasing number of series and reserve the final 28 days as a common holdout. Subsets are drawn by seeded sampling proportional in product category and store, so each approximates the composition of the complete panel while remaining nested across sizes. The sampling seed is fixed across engines and repetitions. The subsets profile resources only; comparative accuracy is reported on the complete panel. We measure data preparation, fitting, prediction, and evaluation separately, recording wall-clock time and peak memory as aggregate proportional set size (PSS) over parent and worker processes, so pages shared between forked processes are counted once. Because the statistical engine can fit several models in one call, we profile both the SeasonalNaive--AutoETS portfolio and AutoETS alone, using the single-model configuration wherever engines are compared with each other, and vary the worker count at two panel sizes so that the parallelization effect can be measured. The evaluation phase times a code path that is identical across engines (the merge of the engine's forecasts with the common holdout, followed by one evaluation call); it is measured with the engine's fitted state still resident, so small per-engine differences in this phase reflect the measurement context rather than engine-specific evaluation code. Thread-count environment variables for the numerical libraries are left unset and recorded per run; parallel configurations may therefore include thread oversubscription between worker processes and the threaded libraries they call. The artifact records the remaining per-run details like software versions, thread settings, device, seed, and CUDA memory for the training process.

The panel-size sweep uses five measured repetitions per configuration and the worker-count sweep three, each in a fresh process so cold-start costs are included and memory is not contaminated by earlier models; we report medians and dispersion. Where two engine configurations are compared with each other rather than across panel sizes, their repetitions are interleaved and their order counterbalanced, so drift in machine state over a long run sequence cannot be mistaken for a difference between configurations. The reconciliation and distributed-backend experiments report one measured run per configuration, because a single configuration takes tens of minutes and, for the dense reconcilers, exhausts machine memory; Sections~\ref{sec:scaling} and~\ref{sec:hierarchy} state the repetitions behind every figure and, where a second pass exists, how far it differs. 

\subsection{Hierarchical workflow}
\label{app:hierarchical}

For the hierarchical dataset, we create base forecasts with AutoETS and LightGBM, store both sets of forecasts in the same keyed dataframe, and reconcile them using fixed BottomUp and MinT configurations where feasible. The MinT configuration is \texttt{MinTraceSparse} with \texttt{method="wls\_var"}: a diagonal approximation of the forecast-error variance, estimated from the in-sample residuals of each base engine, with no non-negativity constraint imposed \citep[for which see][]{wickramasuriya2020nonnegative}. The experiment uses a single 28-day holdout at the final origin rather than the three rolling origins of the evaluation use case. We report MASE by hierarchy level for the base and reconciled forecasts, the reconciliation runtime, and the maximum aggregation residual; the level-wise results diagnose the effect of reconciliation without weighting large revenue series more heavily, whereas WRMSSE in Section~\ref{sec:accuracy} uses the complete official hierarchy and competition weights.

\subsection{External-engine setup}
\label{app:external}

The extensibility claim of Section~\ref{sec:nixtlaverse}, that downstream components accept any engine returning the required keys, cannot be tested with engines built alongside those components. We therefore repeat both downstream tasks with a foreign producer: Croston's method \citep{croston1972forecasting} as implemented in \texttt{sktime} 1.1.0 \citep{loning2019sktime}, an intermittent-demand method appropriate for M5 from an external codebase. The external engine runs in a separate virtual environment with no Nixtla package installed, so the only objects that cross the boundary are the long panel going out and a keyed forecast dataframe coming back. Its forecasts at the same three origins enter the same single evaluation call as the resident engines (Section~\ref{sec:evaluation}), and its base forecasts over the complete 42,840-series hierarchy enter the same sparse bottom-up reconciler (Section~\ref{sec:hierarchy}). Variance-weighted reconciliation is not run for the external engine, because MinT additionally requires in-sample fitted values from the producer.

\subsection{Reproducibility}
\label{app:repro}

The accompanying artifact contains the executable benchmark scripts, direct dependency specification, complete version-pinned dependency closure, M5 file manifest with SHA-256 hashes, model configurations, final raw outcomes, and an evidence map from each manuscript result to its producing command. A second SHA-256 manifest covers the publishable artifact itself. 

Each JSONL outcome records the observed Python and package versions, platform, CPU and memory capacity, thread settings, accelerator, and seed. The reported runs used Python 3.10.12 on Linux/WSL2 with 20 logical CPUs, 48.0~GiB RAM, and an NVIDIA GeForce RTX 5090. Exact runtimes and GPU outcomes remain hardware- and driver-dependent.

\end{document}